\documentclass[10pt,twocolumn,letterpaper]{article}

\usepackage[pagenumbers]{cvpr}

\usepackage{multirow}
\usepackage{array}
\usepackage{tabularx}
\newcolumntype{Y}[1]{>{\hsize=#1\hsize\linewidth=\hsize\centering\arraybackslash}X}

\usepackage{stfloats}

\definecolor{cvprblue}{rgb}{0.21,0.49,0.74}
\usepackage[breaklinks,colorlinks,allcolors=cvprblue]{hyperref}

\def\confName{CVPR}
\def\confYear{2027}

\title{Manga Page Parser: Comprehensive Visual Parsing\\
through Local Story Understanding}

\author{\textit{Yingxuan Li}, \textit{Jiafeng Mao}, \textit{Arata Furukawa}, \textit{Mayu Otani}\\[0.4em]
CyberAgent, Inc.
}

\hypersetup{
  pdftitle={Manga Page Parser: Comprehensive Visual Parsing through Local Story Understanding},
  pdfauthor={Yingxuan Li, Jiafeng Mao, Arata Furukawa, Mayu Otani}
}

\begin{document}
\twocolumn[{
  \maketitle
\begin{minipage}{\textwidth}
    \centering
    \captionsetup{hypcap=false}
    \includegraphics[width=0.90\linewidth]{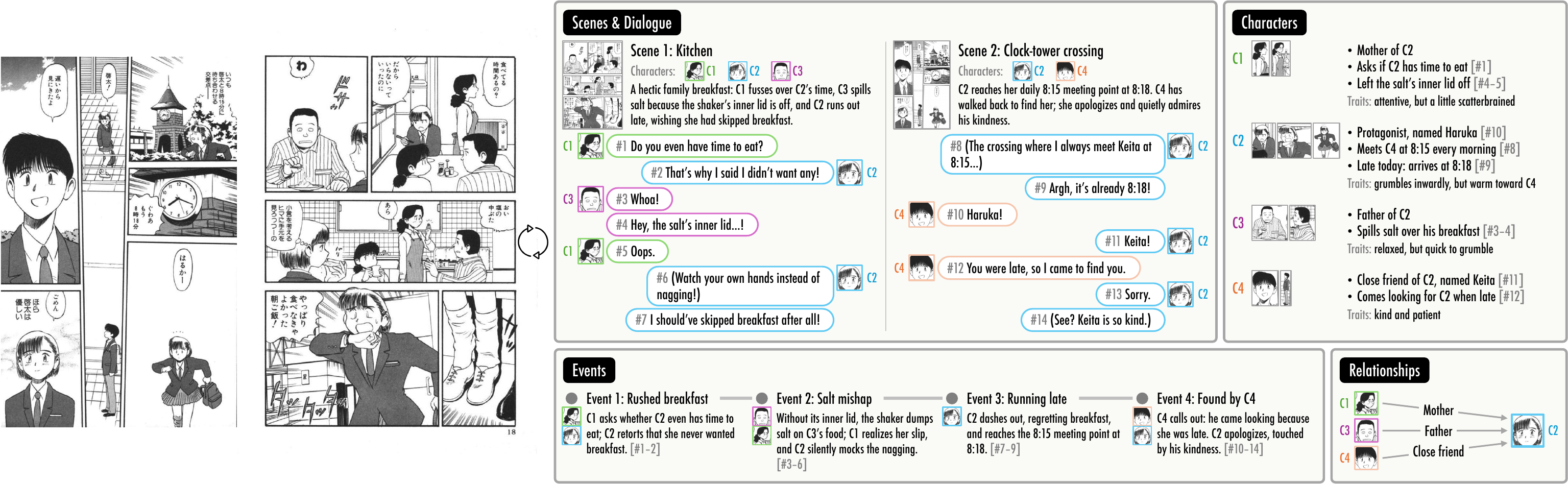}
    \captionof{figure}{Overview of our manga visual parsing task.
    Within each double-page spread, the parser links character appearances and dialogue through shared identities and organizes scenes, events, character states, and relationships into structured records. Courtesy of Ito Shinpei.}
    \label{fig:teaser}
\end{minipage}
\par\medskip

}]
\begin{abstract}
Understanding manga requires connecting who appears and who speaks with the events, character states, and relationships that unfold across panels.
We study manga visual parsing through local story understanding: analyzing each double-page spread independently to capture and organize its story information in a reusable structured representation.
We propose MangaParser, an agentic reading framework that jointly performs character clustering, speaker diarization, and speaker-to-character binding within a shared multimodal conversation, without additional training or fine-tuning.
The agent uses structured story memory to guide character and speaker identification, inspecting relevant image regions when uncertain and refining its predictions and story records together as evidence accumulates.
Quantitative evaluations show that our parser substantially outperforms manga-specific models across multiple backbones.
Parsing examples illustrate how a unified representation links the character and speaker predictions with narrative content and supporting evidence.
We further test reusability through story question answering using outputs generated before the questions are provided.
The results confirm output reusability, offering a foundation for manga translation, content retrieval, and question answering.
\end{abstract}
    
\section{Introduction}
\label{sec:intro}

Manga is a distinctive narrative medium that combines images and dialogue across panels and has gained worldwide popularity.
As the manga market expands and labor costs limit manual processing, demand is growing for automated tools for translation~\cite{lippmann2025context}, content retrieval~\cite{shen2023maru}, question answering~\cite{baek2026mangavqa}, and intellectual property (IP) management.
Supporting these applications requires more than recognizing text and character appearances.
It requires linking dialogue and actions across panels to understand events, characters' motivations, and their relationships.

To meet these needs, we study local story understanding: independently analyzing each double-page spread to capture and structure its story information as comprehensively as possible.
We call this task \textbf{\textit{manga visual parsing through local story understanding}}, coupling character and speaker identification with story interpretation.
Character clustering, speaker diarization, and speaker-to-character binding connect actions and dialogue to characters, helping interpret events, state changes, intentions, and relationships.
In turn, understanding who is present, what characters respond to, and how they interact helps resolve ambiguous identities and speaker assignments.
For example, Figure~\ref{fig:teaser} shows \texttt{Haruka} rushing through breakfast and arriving late to meet \texttt{Keita}.
Tracking her across scenes and identifying the speakers help explain why \texttt{Keita} comes looking for her and why she appreciates his kindness.
The parser preserves these links in structured story records, with character IDs shared within each spread and references to supporting dialogue lines and image regions.
These records can support queries and reasoning across spreads toward understanding the story as a whole.

Previous approaches treat character recognition and speaker prediction as separate tasks~\cite{zhang2023occlusion,li2024manga109dialog} or connect them through predefined processing stages~\cite{li2024zeroshot,sachdeva2025panels}.
However, visual and narrative ambiguities remain challenging.
For example, a speech balloon may be closest to the listener, the same character may change appearance across panels, and a speaker may be outside the current panel.
Resolving these cases requires considering events, dialogue responses, and connections between scenes.
This calls for a shared reading context that supports revising character identities, speaker assignments, and story interpretations together as evidence accumulates.
Moreover, manga-specific recognition models require specialized training, and generalization depends on their training data's coverage of art styles and visual expressions~\cite{zhang2023occlusion,li2024zeroshot,sachdeva2024magi,sachdeva2025panels}.

To address these challenges, we propose MangaParser, an agentic reading framework that jointly reasons about characters and speakers through active reading, without additional training or fine-tuning.
The agent combines images, dialogue, and its evolving understanding of scenes and interactions in a shared multimodal conversation.
Structured story memory preserves this understanding to guide subsequent character and speaker identification.
When uncertain, the agent actively inspects relevant image regions and uses new evidence to refine its predictions and story records together.
This process resembles how human readers combine images and dialogue, revisit details, and refine their understanding.

To validate our parser, we assess its character and speaker identification capabilities, content representation, and downstream reusability.
We quantitatively evaluate speaker diarization, character clustering, and speaker-to-character binding with task-specific metrics.
Our parser substantially outperforms manga-specific models across multiple backbones.
Parsing examples, including Figure~\ref{fig:teaser}, show how the representation connects story content to characters, dialogue, and supporting evidence.
To test reuse, we generate and freeze outputs before any questions are provided, then use them to answer story questions about events, character states, and relationships.
The results demonstrate the value of these outputs as reusable story context for downstream applications.

Our main contributions are as follows:
\begin{itemize}
    \item We formulate manga visual parsing through local story understanding, unifying character identities, dialogue, and narrative content in a structured representation of each double-page spread.
    \item We propose an agentic reading framework combining joint character and speaker reasoning, active image inspection, and structured story memory in a shared multimodal context, without additional training or fine-tuning.
    \item Experiments demonstrate that our method outperforms manga-specific models across multiple backbones and produces reusable parsing outputs for downstream tasks.
\end{itemize}

\begin{figure*}[t]
    \centering
    \includegraphics[width=\textwidth]{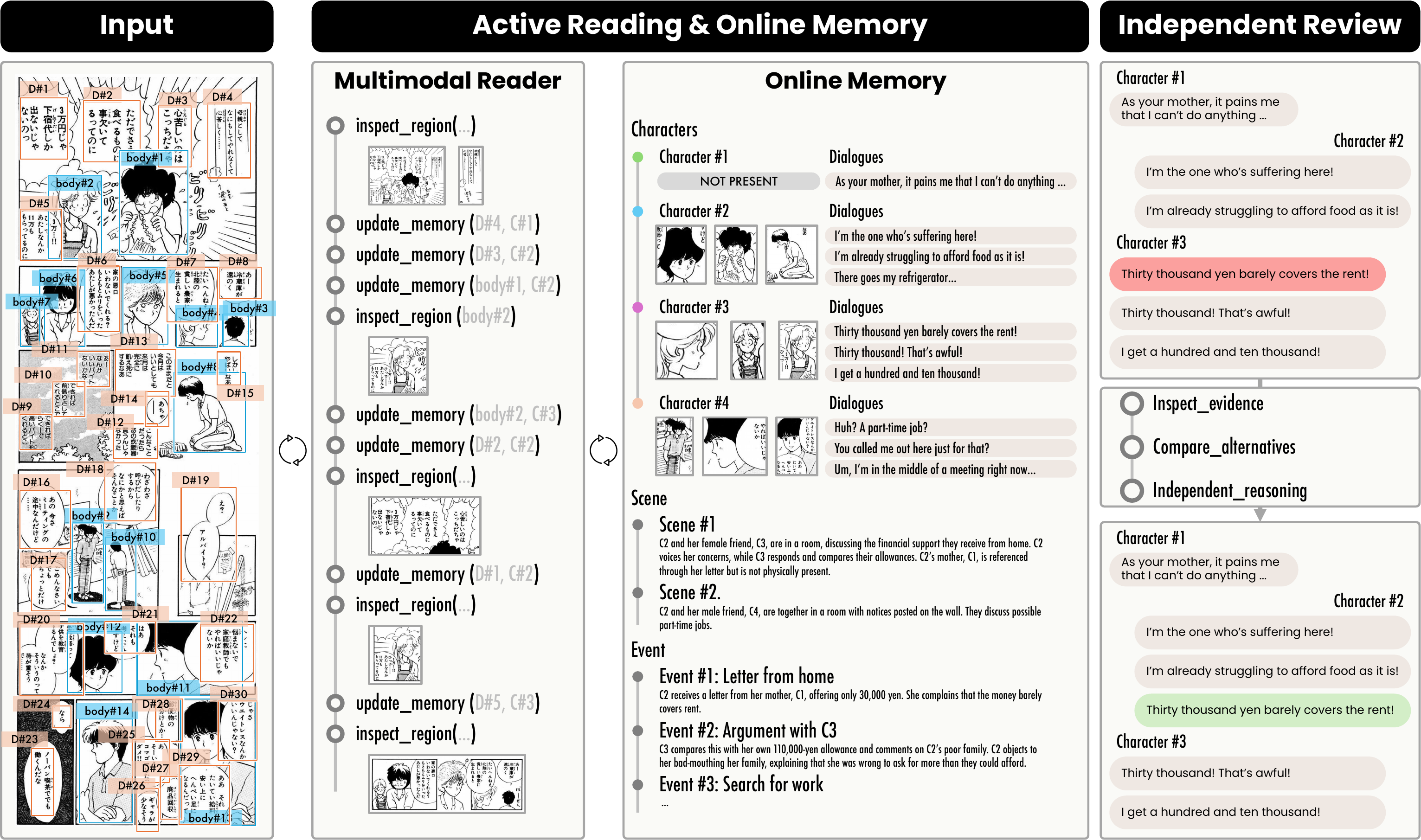}
    \caption{Overview of our agentic manga page parser.
    The Multimodal Reader combines active image inspection with Online Memory to jointly reason about characters, speakers, and story content.
    Optional independent review refines the predictions, after which the story memory is updated.
    Red marks an incorrect speaker attribution; green shows its correction. Courtesy of Yoshi Masako.}
    \label{fig:approach}
\end{figure*}

\section{Related Work}
\label{sec:related_work}

\subsection{Character and Speaker Identification}
Early work treats character and speaker identification as separate tasks~\cite{zhang2023occlusion,li2024manga109dialog}.
Iterative multimodal methods and the MAGI series introduce interactions between these tasks through predefined components and processing stages~\cite{li2024zeroshot,sachdeva2024magi,sachdeva2024tails}.
Their visual recognition components rely on manga-specific training or adaptation.
Our parser instead actively inspects and revisits image regions, combining visual evidence with dialogue and story context without additional training or fine-tuning.

\subsection{Story Understanding and Active Reading}
Beyond identification tasks, manga story understanding has been explored through visual question answering (VQA).
MangaUB evaluates single-panel recognition and multi-panel understanding~\cite{ikuta2025mangaub}, while MangaVQA provides a VQA benchmark with its accompanying specialized MangaLMM model~\cite{baek2026mangavqa}.
For narrative generation, Sachdeva and Zisserman~\cite{sachdeva2025panels} combine MAGI v3 predictions with grounded panel captions to produce prose, placing narrative integration after character and speaker identification.
We instead use story interpretation to refine character and speaker predictions throughout reading.

Active visual understanding also uses tools and memory to gather relevant evidence.
VideoAgent constructs structured video memory and uses tools to retrieve information for a given question~\cite{fan2024videoagent}, while Q-Guide applies question-guided inspection to manga VQA~\cite{popa2026qguide}.
Our parser integrates active image inspection with character and speaker identification, updating structured story memory throughout reading before downstream questions are known.

\section{Approach}
\label{sec:approach}
\label{sec:method}

\subsection{Overview}
\label{sec:overview}

As illustrated in Figure~\ref{fig:approach}, our parser performs local story understanding and organizes the results into a structured representation.
Inputs are the spread image, unordered dialogue transcriptions with line IDs, and character-instance boxes with anonymous IDs.
The parser determines reading order, identifies characters and dialogue speakers, and interprets scenes, events, character states, and relationships.

We provide oracle dialogue transcriptions and character-instance boxes because our focus is on understanding manga and organizing story information.
Optical character recognition (OCR) and element detection in manga have been extensively studied, with dedicated methods achieving strong performance~\cite{hinami2021towards,sachdeva2024magi,sachdeva2025panels}.
Using oracle inputs allows us to evaluate the parser's manga understanding independently of OCR and detection errors.

Given the inputs, the Multimodal Reader actively inspects image regions and maintains Online Memory to jointly reason about character identities, dialogue speakers, and story content.
After reading, an optional Independent Reviewer can check and refine the predictions, triggering updates to the story memory.

\subsection{Active Reading and Online Memory}
\label{sec:active_reading}

As shown in Figure~\ref{fig:approach}, the Multimodal Reader actively inspects image regions using \texttt{inspect\_region}, which returns the crops to its ongoing multimodal conversation.
It combines these visual observations with dialogue and story context to determine reading order and identify characters and dialogue speakers.
When further visual evidence is needed, it can revisit regions of the original image.

The reader maintains Online Memory through \texttt{update\_memory}, recording and refining its predictions and story interpretations throughout reading.
The memory stores reading order, character and speaker predictions, and structured records of scenes, events, character states, relationships, and unresolved questions, such as uncertain dialogue speakers.
Each story record links its content to relevant characters and supporting dialogue or image regions.
When the reader needs additional context during reading, it can retrieve saved story records through \texttt{query\_story}.
The retrieved context supports further reasoning, while new visual or dialogue evidence can revise earlier predictions and story records.

\subsection{Independent Review}
\label{sec:independent_review}

After the Multimodal Reader finishes reading the spread, the Independent Reviewer checks the character and speaker predictions in a separate multimodal conversation.
It receives the original inputs together with these predictions and the predicted reading order.

The reviewer first examines dialogue continuity, question--response relations, and scene context to identify possible inconsistencies in the predictions.
For ambiguous dialogue, it considers alternative interpretations, such as consecutive lines forming one character's monologue or an exchange between speakers.
It compares these alternatives with the current predictions and uses \texttt{inspect\_evidence} to inspect relevant image regions as needed.
In Figure~\ref{fig:approach}, the reviewer checks whether \texttt{C2} or \texttt{C3} speaks the line about 30,000 yen barely covering rent.
The dialogue contrasts the protagonist's financial difficulty with her roommate's larger allowance, supporting the correction to \texttt{C2}.
The reviewer uses \texttt{submit\_decisions} to save corrections together with the original and alternative interpretations, evidence, and conclusion for each issue.

The reviewer revises character-clustering results only with support from dialogue context because manga's stylized depictions and sparse visual cues can make purely visual judgments unreliable.
Each character-instance correction needs both a linked speaker correction and supporting visual evidence.
After review, the system updates the story memory by revising or retiring affected records.

\subsection{Structured Parsing Output}
\label{sec:parsing_output}

After reading and optional review, MangaParser exports reading order, character and speaker predictions, and retained story records.
Shared character IDs connect visual instances, dialogue, and story participants within each spread, with references to supporting dialogue lines and image regions, as shown in Figures~\ref{fig:teaser} and~\ref{fig:approach}.
These outputs provide reusable story context for downstream tasks without repeating parsing.

\section{Main Experiments}
\label{sec:experiments_task123}
\label{sec:experiments}

\begin{figure*}[t]
    \centering
    \includegraphics[width=0.88\textwidth]{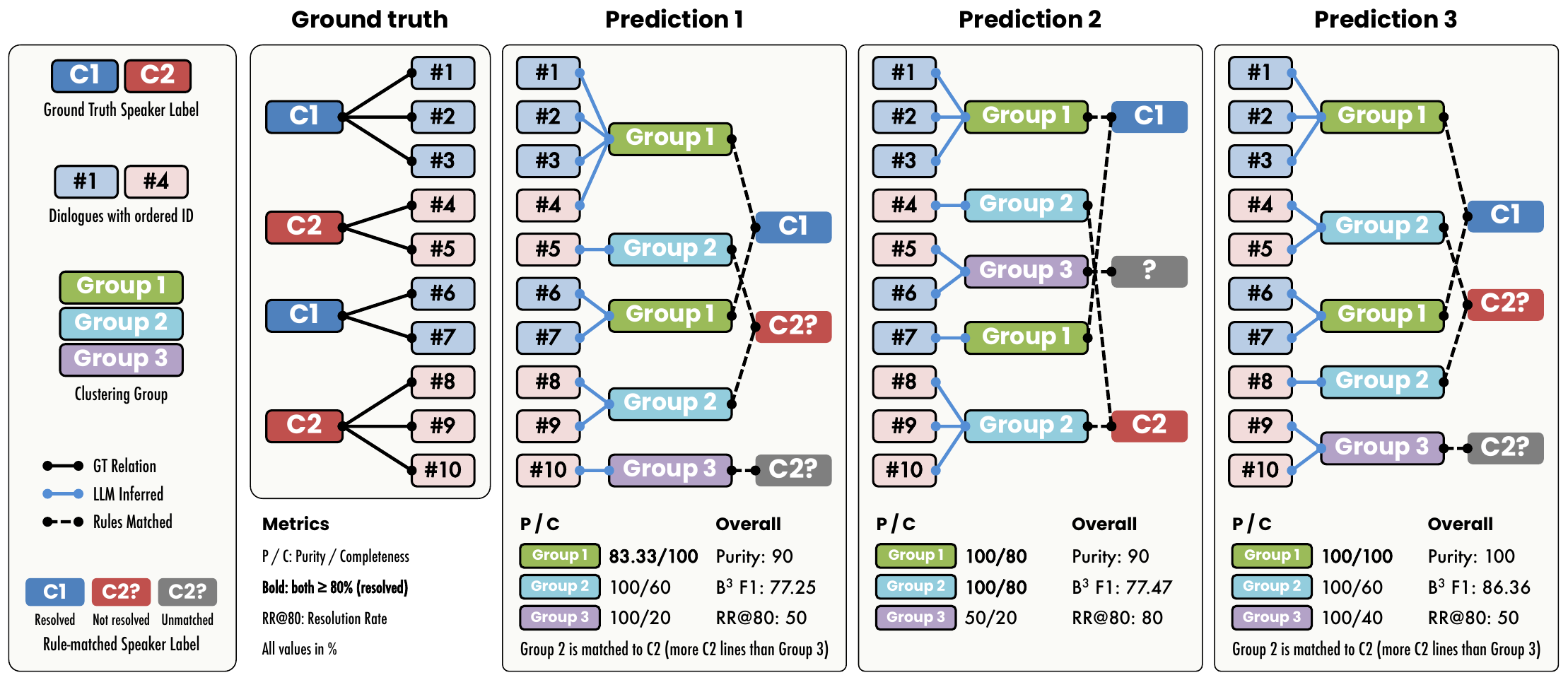}
    \caption{Comparison of conventional clustering metrics and our proposed Resolution Rate (RR). RR@80 counts correct lines in groups with purity and completeness both $\geq80\%$.}
    \label{fig:rr}
\end{figure*}

In our main experiments, we evaluate character and speaker identification through three tasks.

\begin{itemize}
\setlength{\itemsep}{0pt}
\setlength{\parsep}{0pt}
\item \textbf{Task 1: Speaker Diarization.}
Following audio diarization~\cite{fujita2019eend,bredin2023pyannote}, this task groups dialogue lines spoken by the same character without requiring character names or links to visual character instances.
We evaluate this ability explicitly because speaker grouping errors can affect the interpretation of events and interactions.

\item \textbf{Task 2: Character Clustering.}
This task groups visual instances of the same character across panels, evaluating character identification across diverse appearances.

\item \textbf{Task 3: Speaker-to-Character Binding.}
Even with correct speaker grouping, incorrect links to character appearances can change the interpretation of the story.
This task anchors dialogue speakers to the corresponding character instances, evaluating whether dialogue is attributed to the correct characters.
\end{itemize}

\subsection{Experimental Setup}
\label{sec:task123_setup}
\noindent\textbf{Datasets.}
We evaluate the three tasks on Manga109-s, the 87-volume subset of Manga109~\cite{aizawa2020manga109}.
We randomly sample 10 double-page spreads containing dialogue and character instances from each volume, yielding 870 spreads.
The sampled subset contains 11,629 character instances and 10,911 dialogue--speaker annotations from Manga109 and Manga109Dialog~\cite{li2024manga109dialog}, respectively.

\noindent\textbf{Baselines and Setup.}
We compare our framework with a joint one-step large multimodal model (LMM) and manga-specific models, using the state-of-the-art open-source MAGI v2 and v3~\cite{sachdeva2024tails,sachdeva2025panels}.
The one-step baseline jointly predicts characters and dialogue speakers in a single response using the same backbone and inputs as our framework.
MAGI additionally receives oracle text boxes to align its predictions with the evaluation dialogue items.

MAGI's default thresholds leave many dialogue lines without predicted speakers.
We therefore apply threshold-free top-1 decoding to its text-to-character scores: each line is assigned to the character group of its highest-scoring instance, with character clustering unchanged.

\noindent\textbf{Implementation Details.}
We use Qwen3.8-27B~\cite{qwen2026qwen3827b}, Gemini 3.5 Flash~\cite{google2026gemini35flash}, Gemini 3.8 Flash~\cite{google2026gemini38flash}, GPT 5.6 sol~\cite{openai2026gpt56sol}, and GPT 6.1 sol~\cite{openai2026gpt61sol} as backbones for our framework and the joint one-step baseline.
We use medium reasoning effort for character and speaker identification.
We run MAGI v2 and v3 with their publicly released pretrained weights on an NVIDIA A100 GPU (80\,GB).

\subsection{Evaluation Metrics}

We use conventional clustering metrics to evaluate speaker diarization and character clustering, reporting B$^3$ F1~\cite{bagga1998bcubed} for item-level grouping quality and adjusted mutual information (AMI)~\cite{vinh2009ami} for chance-adjusted partition agreement.

For our parsing task, since manga dialogue carries narrative information about characters, mixing speakers can attribute one character's statements to another, while splitting a speaker's dialogue across groups scatters their information across identities.
Under the speaker alignment shown in Figure~\ref{fig:rr}, all three predictions have two assignment errors among ten dialogue lines, but affect character records differently.
Predictions~1 and~2 have similar B$^3$ F1, yet Prediction~1 mixes a \texttt{C2} line into Group~1, potentially confusing the interpretation of \texttt{C1}'s behavior.
Prediction~3 achieves 100\% purity and higher B$^3$ F1, but splits \texttt{C2}'s dialogue between Groups~2 and~3, fragmenting that character's story.

To understand the story more accurately and interpret characters' actions and intentions, we need groups that contain mostly one speaker's dialogue and recover most of that speaker's lines.
B$^3$ F1 and AMI summarize overall clustering quality, but do not report how much dialogue is resolved under both requirements.
We therefore introduce \textbf{\textit{Resolution Rate (RR)}} to measure this coverage under explicit purity and completeness thresholds.

When computing RR, a dialogue line is resolved only if (i) its group's purity and completeness both meet the threshold and (ii) the line is spoken by that group's corresponding ground-truth speaker.
For each predicted group $P$, we choose $G(P)$ as the ground-truth group with the largest overlap; multiple predicted groups may correspond to the same ground-truth group.

The purity and completeness of $P$ are~\cite{manning2008ir,bagga1998bcubed}
\begin{equation}
    p(P)=\frac{|P\cap G(P)|}{|P|},\qquad
    c(P)=\frac{|P\cap G(P)|}{|G(P)|}.
    \label{eq:rr_purity_completeness}
\end{equation}
Let $\mathcal{P}_{\tau}$ contain groups satisfying both $p(P)\geq\tau$ and $c(P)\geq\tau$.
We define
\begin{equation}
    \mathrm{RR}@\tau=\frac{1}{N}\sum_{P\in\mathcal{P}_{\tau}}|P\cap G(P)|,
    \label{eq:resolution_rate}
\end{equation}
where $N$ is the total number of ground-truth dialogue lines (Task 1) or character instances (Task 2).
RR thus counts resolved lines or instances; incorrect items within qualifying groups receive no credit.

In Figure~\ref{fig:rr}, Predictions~1 and 3 each resolve five \texttt{C1} lines in Group~1, while Prediction~2 keeps \texttt{C1} and \texttt{C2} pure and largely complete in Groups~1 and 2, resolving four lines in each group.
The remaining groups in Predictions~1 and 3 fall below the completeness threshold.
Their RR@80 scores are therefore 50\%, 80\%, and 50\%, respectively.

For Tasks 1 and 2, we report RR@80 ($\tau=0.8$), B$^3$ F1, AMI, and RR curves under stricter thresholds.
For Task 3, Binding Accuracy measures the proportion of dialogue lines linked to the correct characters, while Character-macro F1 weights ground-truth speaking characters equally.

\begin{table*}[t]
    \centering
    \caption{Results for Tasks~1--3 on Manga109-s (\%). Spk--Char Binding denotes speaker-to-character binding; Binding Acc. and Macro F1 denote Binding Accuracy and Character-macro F1. $^{\dagger}$ MAGI receives additional oracle text boxes.}
    \label{tab:task123_results}
    \begingroup
    \small
    \setlength{\tabcolsep}{1.75pt}
    \setlength{\aboverulesep}{0pt}
    \setlength{\belowrulesep}{0pt}
    \renewcommand{\arraystretch}{1.3}
    \begin{tabularx}{\textwidth}{@{}c@{\hspace{8pt}}lY{1.02}Y{0.94}Y{0.70}|Y{1.02}Y{0.94}Y{0.70}|Y{1.5}Y{1.18}@{}}
        \toprule
        \rule[-3pt]{0pt}{10pt} &
        & \multicolumn{3}{c|}{\footnotesize\bfseries Task 1: Speaker Diarization}
        & \multicolumn{3}{c|}{\footnotesize\bfseries Task 2: Character Clustering}
        & \multicolumn{2}{c}{\footnotesize\bfseries Task 3: Spk--Char Binding} \\
        Backbone model & Method
        & RR@80 $\uparrow$ & B$^3$ F1 $\uparrow$ & AMI $\uparrow$
        & RR@80 $\uparrow$ & B$^3$ F1 $\uparrow$ & AMI $\uparrow$
        & Binding Acc. $\uparrow$ & Macro F1 $\uparrow$ \\
        \midrule
        \multirow{2}{*}{\shortstack{\textit{Manga-specific}\\\textit{models}}} & MAGI v2$^{\dagger}$~\cite{sachdeva2024tails}
        & 24.48 & 70.06 & 31.50 & 45.83 & 80.77 & 58.51 & 70.91 & 66.04 \\
         & MAGI v3$^{\dagger}$~\cite{sachdeva2025panels}
        & 27.26 & 70.38 & 33.14 & 47.47 & 80.34 & 58.46 & 71.98 & 66.34 \\
        \midrule
        \multirow{3}{*}{\textit{Qwen3.8-27B}} & Joint one-step LMM
        & 31.00 & 68.64 & 38.38 & 38.98 & 70.55 & 48.58 & 68.02 & 62.10 \\
         & MangaParser (Ours)
        & 36.69 & 72.23 & 44.00 & 39.56 & 70.46 & 48.59 & 71.89 & 65.53 \\
         & \hspace{1.5em}+ Review (Ours)
        & \textbf{39.04} & \textbf{72.96} & \textbf{45.68} & \textbf{39.94} & \textbf{70.61} & \textbf{48.85} & \textbf{72.64} & \textbf{66.70} \\
        \midrule
        \multirow{3}{*}{\textit{Gemini 3.5 Flash}} & Joint one-step LMM
        & 49.85 & 77.05 & 55.33 & 17.86 & 54.66 & 19.02 & 57.37 & 48.84 \\
         & MangaParser (Ours)
        & 63.92 & 84.04 & 68.22 & \textbf{21.22} & \textbf{57.40} & \textbf{24.99} & \textbf{63.58} & \textbf{54.40} \\
         & \hspace{1.5em}+ Review (Ours)
        & \textbf{66.67} & \textbf{84.60} & \textbf{69.59} & 20.58 & 56.93 & 23.90 & 62.65 & 53.55 \\
        \midrule
        \multirow{3}{*}{\textit{Gemini 3.8 Flash}} & Joint one-step LMM
        & 59.00 & 81.73 & 64.28 & 36.29 & 65.88 & 42.17 & 74.20 & 68.37 \\
         & MangaParser (Ours)
        & 75.69 & 88.71 & 77.38 & \textbf{71.25} & \textbf{85.75} & \textbf{74.07} & 86.62 & 83.17 \\
         & \hspace{1.5em}+ Review (Ours)
        & \textbf{77.94} & \textbf{89.67} & \textbf{79.57} & 71.11 & 85.63 & 73.83 & \textbf{87.06} & \textbf{83.64} \\
        \midrule
        \multirow{3}{*}{\textit{GPT 5.6 sol}} & Joint one-step LMM
        & 63.04 & 83.94 & 66.41 & 65.50 & 84.85 & 71.09 & 82.74 & 77.93 \\
         & MangaParser (Ours)
        & 68.80 & 86.05 & 71.26 & 72.41 & 88.08 & 77.66 & 85.66 & 81.90 \\
         & \hspace{1.5em}+ Review (Ours)
        & \textbf{68.91} & \textbf{86.26} & \textbf{71.70} & \textbf{72.47} & \textbf{88.12} & \textbf{77.73} & \textbf{86.12} & \textbf{82.33} \\
        \midrule
        \multirow{3}{*}{\textit{GPT 6.1 sol}} & Joint one-step LMM
        & 79.13 & 90.16 & 80.30 & 83.79 & 93.06 & 87.37 & 90.42 & 87.45 \\
         & MangaParser (Ours)
        & \textbf{83.39} & \textbf{91.98} & \textbf{83.69} & 86.95 & 94.42 & 90.00 & \textbf{92.53} & \textbf{90.44} \\
         & \hspace{1.5em}+ Review (Ours)
        & 83.19 & 91.91 & 83.54 & \textbf{87.02} & \textbf{94.46} & \textbf{90.10} & 92.48 & 90.22 \\
        \bottomrule
    \end{tabularx}
    \endgroup
\end{table*}

\subsection{Main Results}
\label{sec:main_results}
Table~\ref{tab:task123_results} reports results on the three character and speaker identification tasks.
MangaParser denotes the active reader with Online Memory; MangaParser + Review adds independent review.

For speaker diarization, LMM-based methods generally outperform the MAGI series, with competitive results even from the open-source Qwen3.8-27B.
The joint one-step LMM trails MAGI in character clustering on three backbones, with Qwen3.8-27B and Gemini 3.5 Flash also showing lower speaker-to-character binding accuracy.
This pattern is consistent with the difficulty of distinguishing manga characters from stylized appearances and sparse visual cues.

Our agentic framework addresses these visual recognition difficulties.
With Gemini 3.8 Flash, character clustering RR@80 increases from 36.29\% to 71.25\%, exceeding MAGI v3's 47.47\% without additional training.
Binding Accuracy rises from 74.20\% to 86.62\%.
These gains support combining visual evidence with dialogue and local story context for character and speaker identification.

\begin{figure}[!t]
    \centering
    \includegraphics[width=0.92\linewidth]{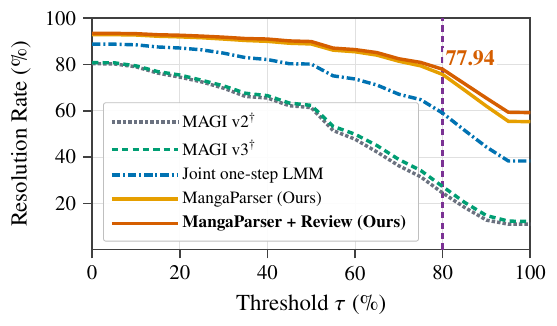}
    \caption{Speaker diarization RR. LMM-based methods use Gemini 3.8 Flash. $^{\dagger}$ MAGI receives additional oracle text boxes.}
    \label{fig:rr_curves}
    \label{fig:rr_curve_speaker}
\end{figure}

Independent review further improves speaker diarization RR@80 on four of the five backbones.
With Gemini 3.8 Flash, it raises RR@80 from 75.69\% to 77.94\%.
As Figure~\ref{fig:rr_curves} also shows, this benefit persists even at the strictest threshold of 100\%, where MangaParser + Review achieves an RR of 59.19\%.
For speaker-to-character binding with the same backbone, MangaParser + Review reaches at least 80\% accuracy on the sampled spreads in 75 of 87 volumes.
Independent review also improves speaker diarization RR@80 in 52 volumes.
GPT 6.1 sol achieves the highest overall scores, highlighting the potential of advances in LMMs for manga understanding.

\subsection{Ablation Study}
\label{sec:ablation}
\newlength{\qualitativeimageheight}
\begin{figure*}[t]
    \centering
    \setlength{\qualitativeimageheight}{0.36\textwidth}
    \begin{subfigure}[b]{0.49\textwidth}
        \centering
        \includegraphics[height=\qualitativeimageheight]{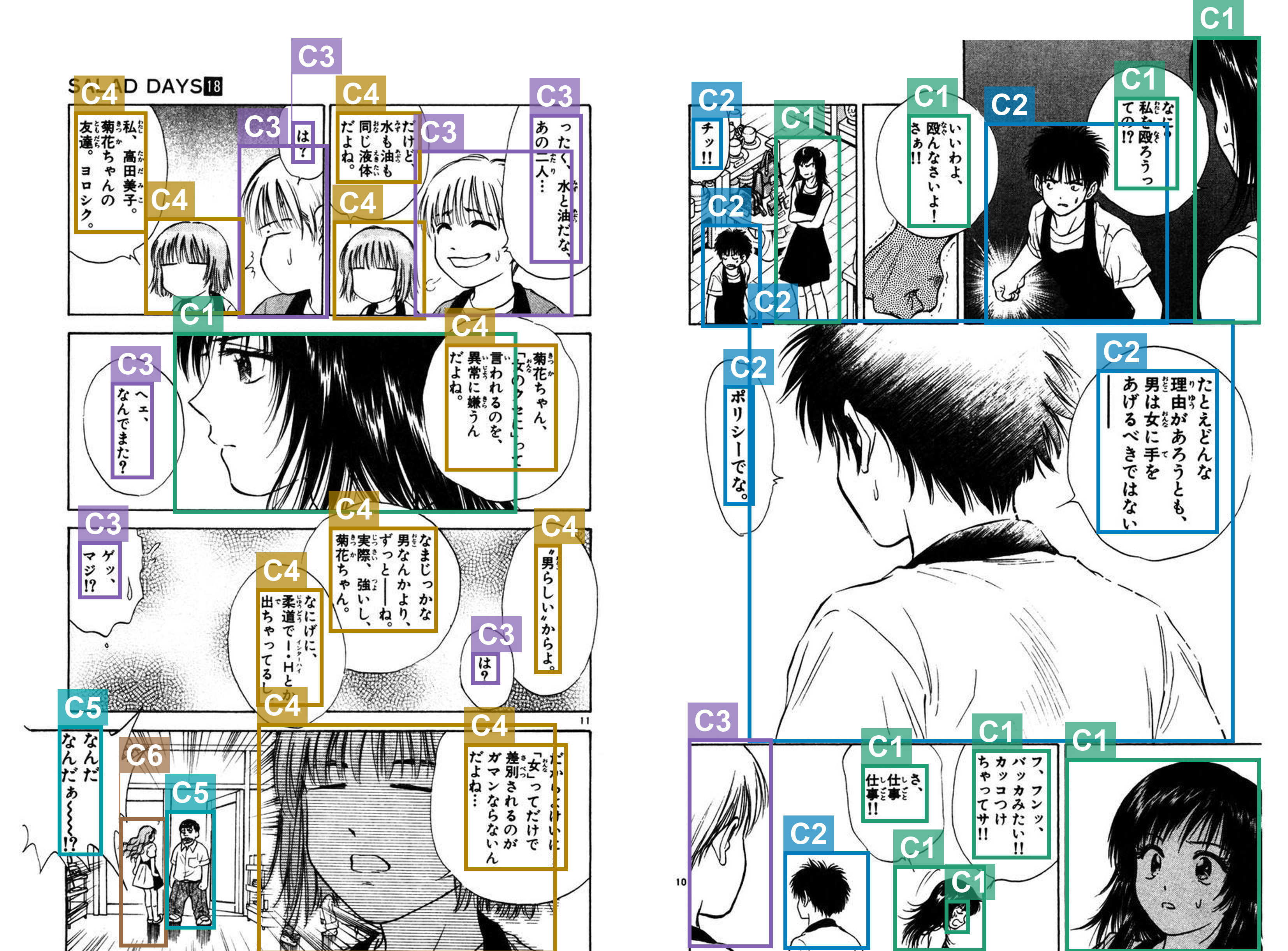}
        \caption{Courtesy of Inokuma Shinobu.}
    \end{subfigure}\hfill
    \begin{subfigure}[b]{0.49\textwidth}
        \centering
        \includegraphics[height=\qualitativeimageheight]{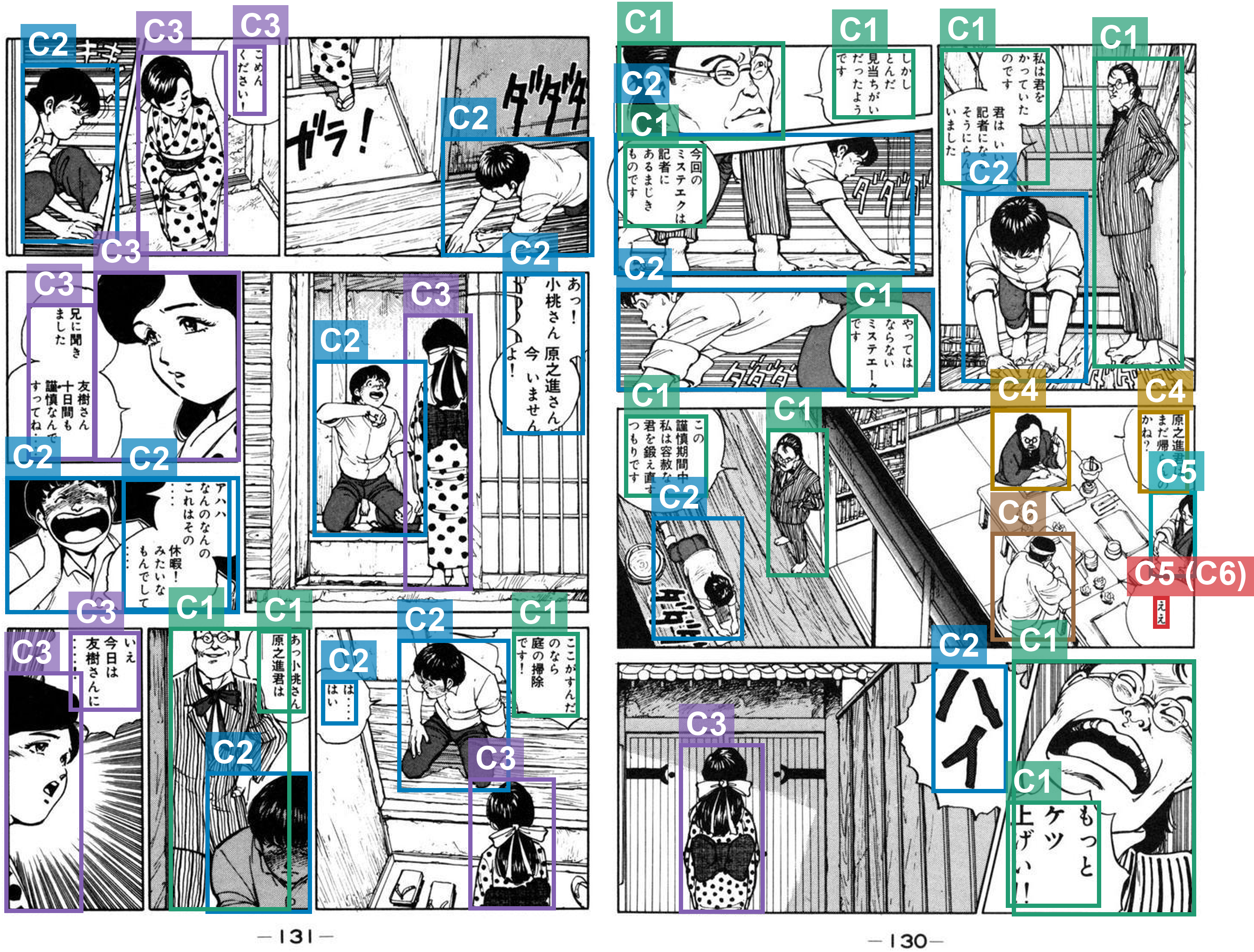}
        \caption{Courtesy of Kimura Naomi.}
    \end{subfigure}
    \caption{Examples of character and speaker identification from our MangaParser using Gemini 3.8 Flash. Shared colors and character IDs link character instances and dialogue within each spread. Red marks incorrect predictions; IDs in parentheses indicate ground truth.}
    \label{fig:qualitative_examples}
\end{figure*}

We compare static memory and online memory for our framework on one manga title (94 double-page spreads with 1,337 dialogue lines).
Static memory is generated after reasoning, whereas online memory supports updates and queries during reasoning.
We also compare online and post-hoc review starting from the same initial predictions.
Table~\ref{tab:ablation} reports the speaker diarization results using Gemini 3.8 Flash with the same seed.
The results show higher performance with online memory, which provides relevant story context during reasoning.
We adopt post-hoc review because it achieves higher RR@80 and B$^3$ F1 at slightly lower cost than online review, while its separation from the reader simplifies controlled experiments.

\begin{table}[t]
    \centering
    \caption{Speaker diarization ablation on one manga title using Gemini 3.8 Flash (\%). Token cost is relative to Online Memory.}
    \label{tab:ablation}
    \begingroup
    \small
    \setlength{\tabcolsep}{1.5pt}
    \renewcommand{\arraystretch}{1.2}
    \begin{tabular*}{\columnwidth}{@{\extracolsep{\fill}}lcccc@{}}
        \toprule
        Method
        & RR@80 $\uparrow$
        & B$^3$ F1 $\uparrow$
        & AMI $\uparrow$
        & Rel. Cost $\downarrow$ \\
        \midrule
        Static memory
        & 74.20 & 88.60 & 73.05 & $0.70\times$ \\
        Online Memory
        & 79.73 & 90.64 & 78.40 & $1.00\times$ \\
        \hspace{1.5em}+ Online Reviewer
        & 82.72 & 92.41 & \textbf{83.13} & $1.36\times$ \\
        \hspace{1.5em}+ Post-hoc Reviewer
        & \textbf{84.89} & \textbf{92.58} & 82.65 & $1.33\times$ \\
        \bottomrule
    \end{tabular*}
    \endgroup
\end{table}

\subsection{Qualitative Evaluation}
\label{sec:qualitative_evaluation}

Figure~\ref{fig:qualitative_examples} shows character and speaker identification examples from our MangaParser using Gemini 3.8 Flash.
Shared colors and character IDs link visual instances and dialogue, while red marks incorrect predictions.
Overall, our parser effectively identifies characters and dialogue speakers, tracking character entrances and exits across panels.
For speaker prediction, it correctly attributes the alternating dialogue of \texttt{C3} and \texttt{C4} on the left page in Figure~\ref{fig:qualitative_examples}~(a), even when they are absent from the current panel.
In Figure~\ref{fig:qualitative_examples}~(b), it also correctly identifies speakers who are not closest to the speech balloon.
The only speaker error misattributes a background character's reply.
Here, the dialogue and balloon direction are ambiguous, and the questioner's facial orientation provides the key cue.
This illustrates a challenging case for our current method.

For character identification, all visual instances are correctly grouped in both examples.
The parser recognizes characters despite substantial deformations, such as \texttt{C4} in Figure~\ref{fig:qualitative_examples}~(a) and \texttt{C2} in Figure~\ref{fig:qualitative_examples}~(b).
It also handles changes in shot scale and viewpoint, such as \texttt{C1} in Figure~\ref{fig:qualitative_examples}~(a) and \texttt{C3} in Figure~\ref{fig:qualitative_examples}~(b).
These examples illustrate the value of combining visual evidence with local story context for character and speaker identification.
Additional examples are provided in the supplementary material.

\section{Experiments on Story Question Answering}
\label{sec:story_qa}

In addition to the main experiments, we evaluate the reusability of our parser's outputs through story question answering (Story QA).

\subsection{Experimental Setup}
\noindent\textbf{Datasets.}
We evaluate Story QA using MangaVQA~\cite{baek2026mangavqa}, a benchmark containing 526 manually constructed question--answer pairs for contextual manga understanding.
From this benchmark, we select 225 questions from 205 double-page spreads across eight manga volumes.
We focus on questions about characters, events, states, relationships, motivations, and causal explanations, excluding those that require only reading or copying text.

\noindent\textbf{Baselines and Setup.}
We evaluate two modes of output reuse: (i) answering story questions from parsing outputs generated before questions are known; (ii) using these outputs as memory to improve image-based QA.
All parsers process the original images using the oracle setup in Section~\ref{sec:overview}, and their outputs remain fixed throughout QA.

A shared text-only reader answers questions solely from the outputs of MAGI v3, the joint one-step LMM, and MangaParser.
The joint one-step baseline generates static memory from the image, oracle inputs, and fixed character and speaker predictions, using the same structured format as MangaParser's story memory.
For MAGI v3, we follow its original pipeline of panel captioning, character grounding, and narrative generation~\cite{sachdeva2025panels}.
We use the outputs of this original pipeline, including narrative prose, panel descriptions, and character and speaker predictions, for QA.

To test memory augmentation, we evaluate both MangaLMM~\cite{baek2026mangavqa} and LMM-based VQA with and without MangaParser memory.
All receive the original spread, question, oracle dialogue transcriptions, and anonymous character-instance boxes.
For QA with memory, we provide MangaParser's story memory.
For LMM-based VQA, parsing and answering use the same backbone; for MangaLMM, we use its publicly released pretrained model.

\begin{table*}[t]
    \centering
    \caption{Story QA accuracy on 225 MangaVQA questions (\%). $^{\dagger}$ MAGI receives additional oracle text boxes. $^{\ddagger}$ The joint one-step LMM generates static memory after its character and speaker predictions are fixed. MangaLMM uses the same pretrained model across columns.}
    \label{tab:story_qa_results}
    \begingroup
    \small
    \setlength{\tabcolsep}{1pt}
    \setlength{\aboverulesep}{0pt}
    \setlength{\belowrulesep}{0pt}
    \renewcommand{\arraystretch}{1.3}
    \begin{tabularx}{\textwidth}{@{}>{\raggedright\arraybackslash}p{80pt}>{\raggedright\arraybackslash}p{102pt}*{5}{>{\centering\arraybackslash}X}@{}}
        \toprule
        \rule[-4pt]{0pt}{10pt} & Method
        & \mbox{\footnotesize\textit{Qwen3.8-27B}}
        & \mbox{\footnotesize\textit{Gemini 3.5 Flash}}
        & \mbox{\footnotesize\textit{Gemini 3.8 Flash}}
        & \mbox{\footnotesize\textit{GPT 5.6 sol}}
        & \mbox{\footnotesize\textit{GPT 6.1 sol}} \\
        \midrule
        \multirow{3}{*}{\shortstack[l]{w/o original\\image at QA time}}
        & MAGI v3$^{\dagger}$~\cite{sachdeva2025panels}
        & \textbf{85.33} & 88.44 & 88.89 & 85.33 & 85.33 \\
        & Joint one-step LMM$^{\ddagger}$
        & 75.56 & \textbf{89.33} & 88.44 & 84.44 & 89.78 \\
        & MangaParser (Ours)
        & 81.78 & 88.00 & \textbf{92.44} & \textbf{86.67} & \textbf{93.78} \\
        \midrule
        \multirow{4}{*}{\shortstack[l]{w/ original\\image at QA time}}
        & MangaLMM~\cite{baek2026mangavqa}
        & 49.33 & 49.33 & 49.33 & 49.33 & 49.33 \\
        & \hspace*{1em}+ MangaParser memory
        & 48.89 & 53.33 & 59.11 & 59.11 & 57.33 \\
        & LMM-based VQA
        & 79.56 & 90.22 & 95.56 & 89.78 & 96.00 \\
        & \hspace*{1em}+ MangaParser memory
        & \textbf{82.22} & \textbf{92.44} & \textbf{97.33} & \textbf{92.89} & \textbf{96.89} \\
        \bottomrule
    \end{tabularx}
    \endgroup
\end{table*}

\noindent\textbf{Evaluation.}
All answers are evaluated through a shared large language model (LLM) judge using the question, reference answer, and original dialogue in MangaParser's predicted reading order.
Uncertain cases are adjudicated manually using the images.
We report overall accuracy on all 225 questions, marking answers correct only when they accurately and completely convey the requested information.
Empty answers, refusals, and method failures are counted as incorrect.

\noindent\textbf{Implementation Details.}
We use the backbones listed in Table~\ref{tab:story_qa_results} for LMM-based parsing and VQA, including MAGI v3's captioning and narrative generation.
Gemini 3.8 Flash serves as the shared text-only QA reader and LLM judge.
We use medium reasoning effort for the shared QA reader, LMM-based VQA, and the judge.
Local inference uses NVIDIA A100 GPUs (80\,GB).

\subsection{Results}
\label{sec:story_qa_results}
Table~\ref{tab:story_qa_results} reports results for answering questions from parsing outputs and using these outputs as QA memory.

\begin{figure}[!t]
    \centering
    \includegraphics[width=\columnwidth]{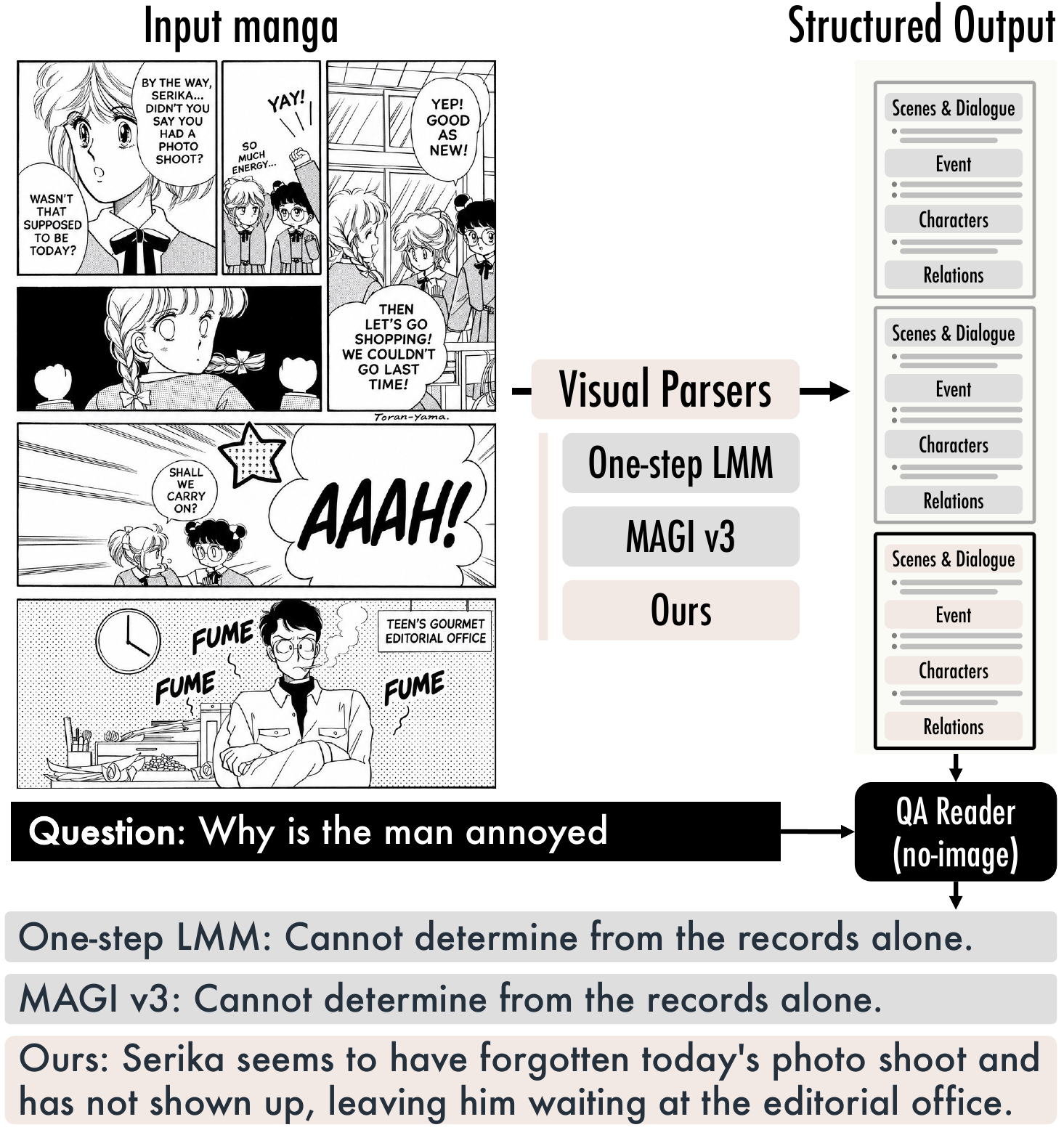}
    \caption{Story QA from GPT 5.6 sol and 6.1 sol parsing outputs. The original image and QA text are in Japanese and have been translated into English for presentation. Courtesy of Kurita Riku.}
    \label{fig:story_qa_qualitative}
\end{figure}

For QA without images, MangaParser outperforms the joint one-step LMM on four of the five backbones.
It also outperforms MAGI v3 with Gemini 3.8 Flash and both GPT backbones.
An example is shown in Figure~\ref{fig:story_qa_qualitative}.
No dialogue directly explains why the man is annoyed, so answering requires understanding the story on this page.
\texttt{Serika} prepares to go shopping, freezes when a friend asks whether her photo shoot is today, and then cries out.
The scene then shifts to a man waiting impatiently at the editorial office.
Answering requires connecting the shopping conversation, her reaction, and his waiting to infer that she has forgotten the appointment.
The illustrated results use GPT backbones: only MangaParser's outputs support the correct answer, while the joint one-step LMM and MAGI v3 fail to answer correctly.
These results show that our parser provides rich information to support local story understanding.

For image-based QA, MangaParser memory improves LMM-based VQA across all five backbones.
With Gemini 3.8 Flash, accuracy rises from 95.56\% to 97.33\%, showing that the stored story information provides useful context alongside the original images.
MangaLMM also benefits from memory generated by the Gemini and GPT backbones, with accuracy increasing from 49.33\% to 53.33--59.11\%.

The results in Table~\ref{tab:story_qa_results} show that MangaParser's outputs effectively support Story QA, highlighting the potential of our structured story memory for applications such as content retrieval and manga translation.
The evaluated closed-source backbones outperform the open-source Qwen3.8-27B, suggesting stronger manga understanding capabilities in this setting.
We provide QA results by question type and additional QA examples in the supplementary material.

\section{Conclusions}
\label{sec:conclusion}
We presented MangaParser, an agentic manga page parser for comprehensive visual parsing through local story understanding.
By combining active image inspection with structured story memory, it jointly identifies characters and speakers and organizes each double-page spread into reusable story records without additional training or fine-tuning.
These records retain explicit links to supporting evidence.
Experiments across multiple backbones demonstrate substantial improvements over manga-specific models.
Story QA confirms that the parsing outputs can support question answering and serve as memory to enhance QA.
Although challenging cases remain, these findings demonstrate the value of comprehensive visual parsing for manga understanding, providing a foundation for downstream applications such as manga translation.

\section*{LLM Usage Statement}
We used LLM-based tools, including ChatGPT and Codex, to assist with English language editing, implementation of author-designed methods and experiments, analysis of experimental results and error cases, and preparation and refinement of figures and diagrams.
These tools were also used to translate manga dialogue and related text into English for presentation in figures.
The authors reviewed and verified all AI-assisted outputs and take full responsibility for the research ideas, methodological decisions, experimental designs, interpretation of results, and final manuscript.

{\small\bibliographystyle{ieeenat_fullname}\bibliography{main}}

\clearpage
\appendix
\setcounter{figure}{0}
\setcounter{table}{0}
\renewcommand{\thefigure}{\Alph{figure}}
\renewcommand{\thetable}{\Alph{table}}
\renewcommand{\theHfigure}{supp.\Alph{figure}}
\renewcommand{\theHtable}{supp.\Alph{table}}
\maketitlesupplementary
In this supplementary material, we present additional examples of character and speaker identification and story question answering (Story QA).
We also report and analyze Story QA results by question type.

\newlength{\suppqualitativeimageheight}
\begin{figure*}[!b]
    \centering
    \setlength{\suppqualitativeimageheight}{0.345\textwidth}
    \begin{subfigure}[b]{0.49\textwidth}
        \centering
        \includegraphics[height=\suppqualitativeimageheight]{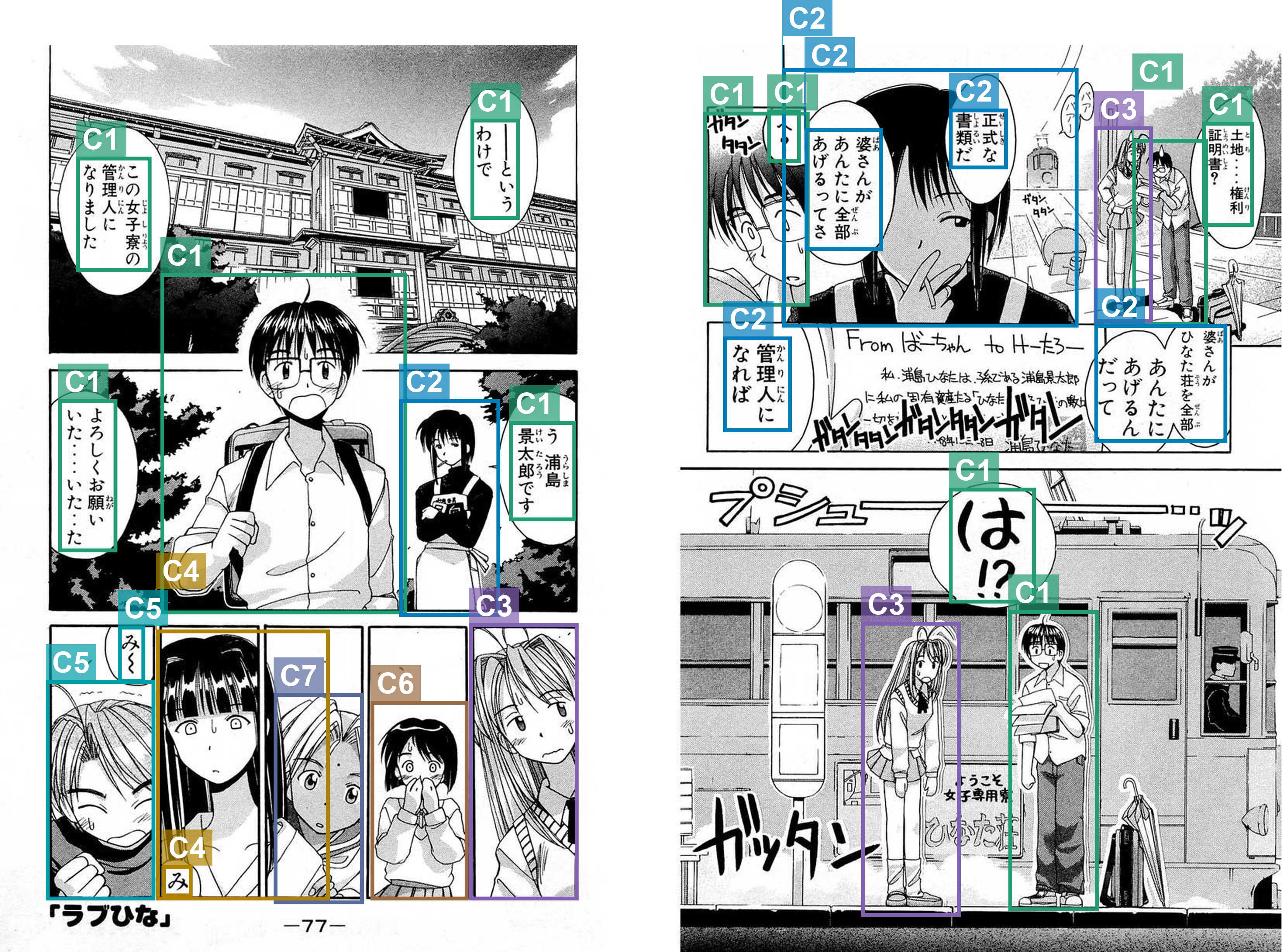}
        \caption{Courtesy of Akamatsu Ken.}
        \label{fig:supp_lovehina}
    \end{subfigure}\hfill
    \begin{subfigure}[b]{0.49\textwidth}
        \centering
        \includegraphics[height=\suppqualitativeimageheight]{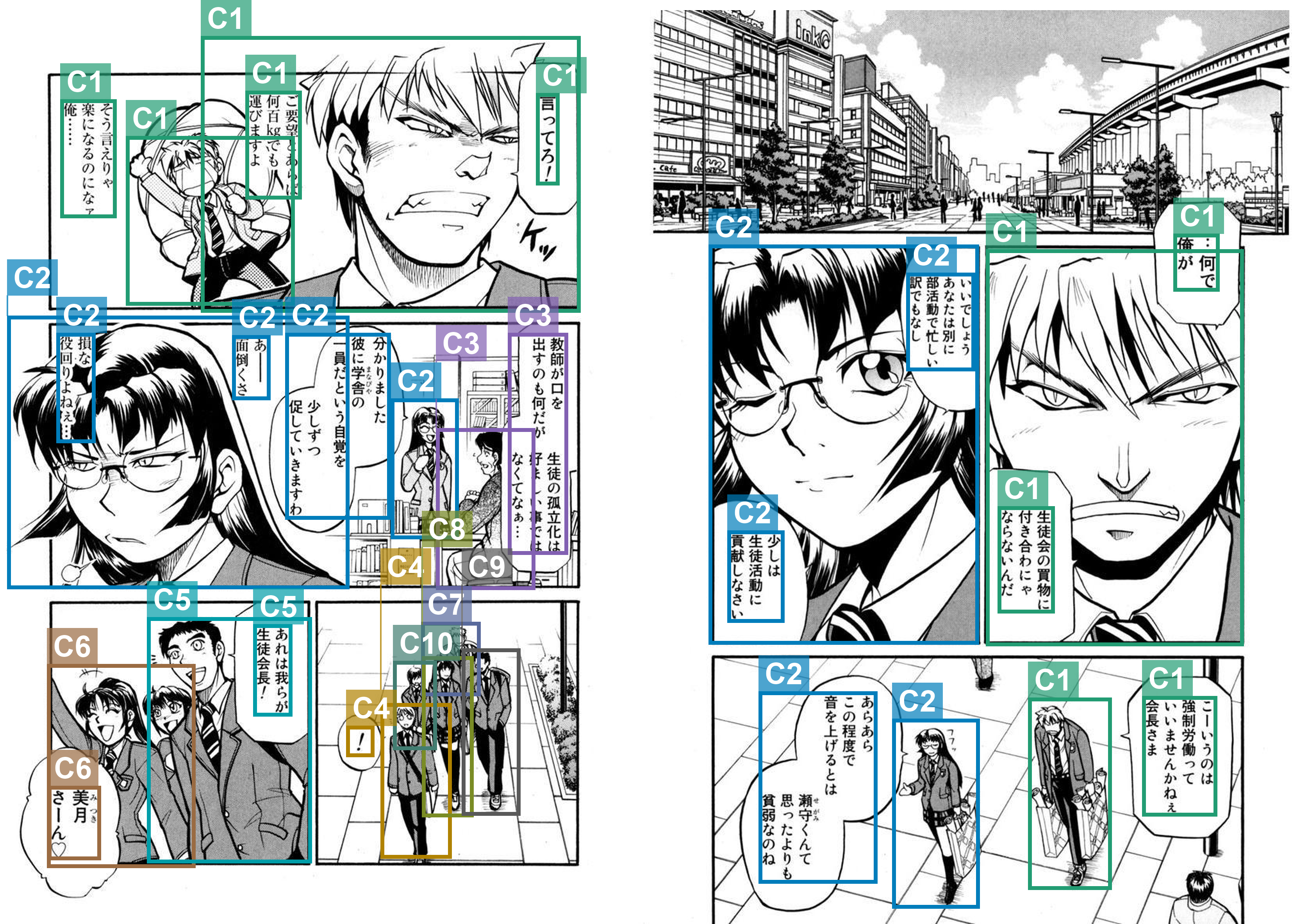}
        \caption{Courtesy of Takeyama Yusuke.}
        \label{fig:supp_dualjustice}
    \end{subfigure}

    \medskip
    \begin{subfigure}[b]{0.49\textwidth}
        \centering
        \includegraphics[height=\suppqualitativeimageheight]{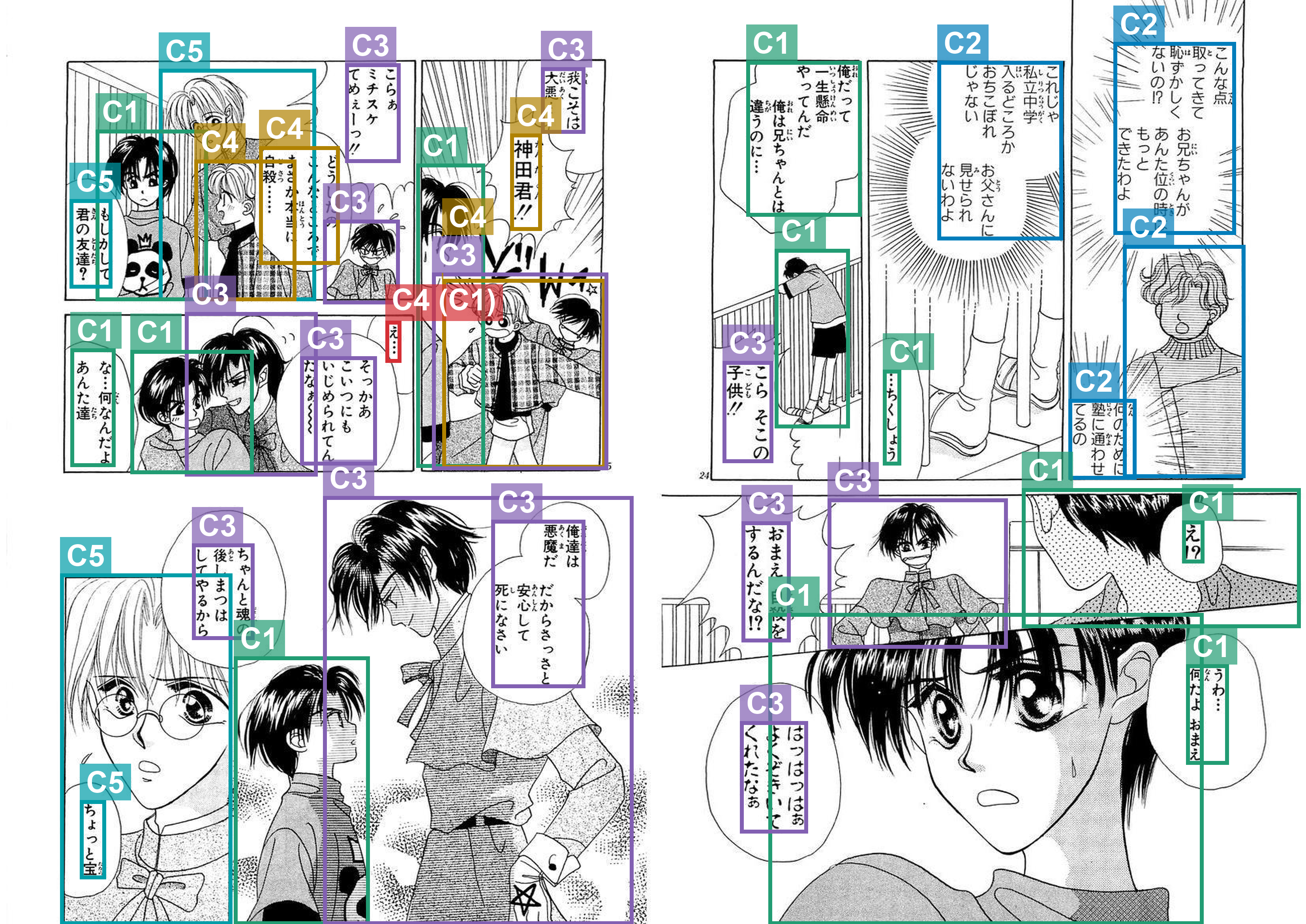}
        \caption{Courtesy of Hanada Sakumi.}
        \label{fig:supp_yasasiiakuma}
    \end{subfigure}\hfill
    \begin{subfigure}[b]{0.49\textwidth}
        \centering
        \includegraphics[height=\suppqualitativeimageheight]{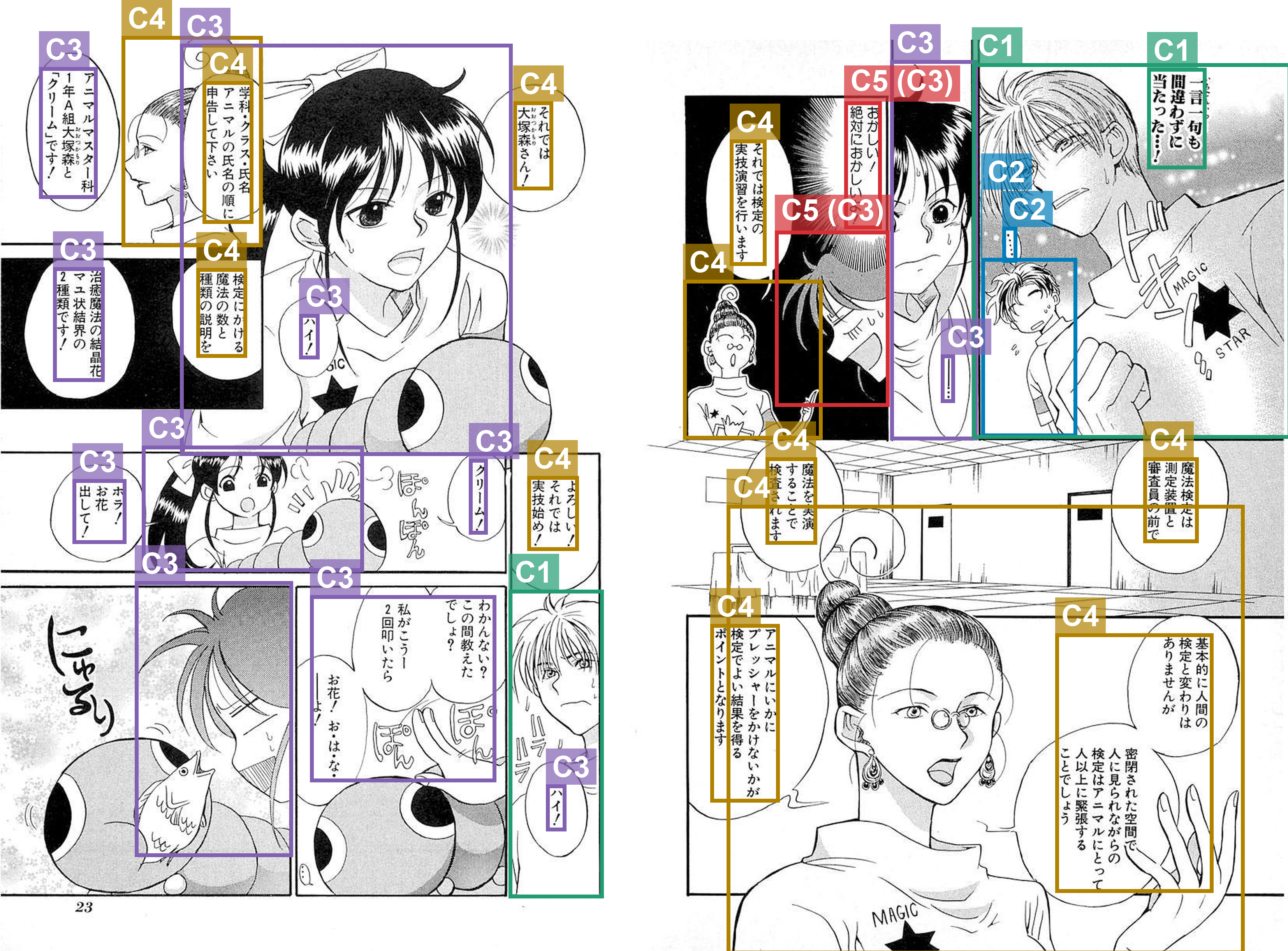}
        \caption{Courtesy of Minamizawa Hisaka.}
        \label{fig:supp_magicstargakuin}
    \end{subfigure}
    \caption{Examples of character and speaker identification from MangaParser using Gemini 3.8 Flash.
    Shared colors and character IDs link visual instances and dialogue within each spread.
    Red marks incorrect predictions; IDs in parentheses indicate ground truth.}
    \label{fig:supp_qualitative_examples}
\end{figure*}

\section{Character and Speaker Identification Examples}
\label{sec:supp_task123_examples}

Figure~\ref{fig:supp_qualitative_examples} provides additional examples of character clustering, speaker diarization, and speaker-to-character binding from MangaParser using Gemini 3.8 Flash.
As in the main paper, shared colors and character IDs link visual instances and dialogue, while red marks incorrect predictions.

In the multi-character scenes in Figures~\ref{fig:supp_qualitative_examples}~(a) and~(b), our parser correctly identifies all evaluated character instances and dialogue speakers.
In particular, in Figure~\ref{fig:supp_qualitative_examples}~(b), the parser recognizes \texttt{C1} in both normal and chibi depictions and \texttt{C2} in both close-up and distant views.
It also correctly attributes recalled dialogue and inner speech alongside the ongoing conversation.
These examples illustrate the importance of combining visual evidence with local story context for character and speaker identification.

Cases with limited contextual or visual cues remain challenging.
In Figure~\ref{fig:supp_qualitative_examples}~(c), a brief response with no clear balloon direction is attributed to the wrong speaker.
In Figure~\ref{fig:supp_qualitative_examples}~(d), the same character is identified as different characters due to large differences in drawing style and appearance.
These errors highlight the need for deeper reasoning from sparse visual and narrative evidence.

\section{Story QA Results by Question Type}
\label{sec:supp_qa_by_type}

Table~\ref{tab:supp_story_qa_by_type} reports Story QA accuracy by question type under the Gemini 3.8 Flash setting of the main paper.
All parsing outputs are generated before questions are provided and remain fixed during QA.

MangaVQA categorizes questions by information scope (Panel or Page) and answer type (Exact extraction or Descriptive answering)~\cite{baek2026mangavqa}.
Our 225-question subset contains 83 Panel and 142 Page questions, with 71 Exact extraction and 154 Descriptive answering questions.

\begin{table*}[t]
    \centering
    \caption{Story QA accuracy by question type using Gemini 3.8 Flash on 225 MangaVQA questions (\%).
    Exact extr. denotes Exact extraction, and Descriptive ans. denotes Descriptive answering.
    $^{\dagger}$ MAGI receives additional oracle text boxes.
    $^{\ddagger}$ The joint one-step LMM generates static memory after its character and speaker predictions are fixed.}
    \label{tab:supp_story_qa_by_type}
    \begingroup
    \small
    \setlength{\tabcolsep}{3pt}
    \setlength{\aboverulesep}{0pt}
    \setlength{\belowrulesep}{0pt}
    \renewcommand{\arraystretch}{1.3}
    \begin{tabular}{@{}
        >{\raggedright\arraybackslash}p{72pt}
        >{\raggedright\arraybackslash}p{130pt}
        >{\centering\arraybackslash}p{48pt}|
        *{2}{>{\centering\arraybackslash}p{40pt}}|
        >{\centering\arraybackslash}p{60pt}
        >{\centering\arraybackslash}p{68pt}@{}}
        \toprule
        & & & \multicolumn{2}{c|}{\textbf{Required information}}
        & \multicolumn{2}{c}{\textbf{Answer type}} \\
        & Method & Overall $\uparrow$
        & Panel $\uparrow$ & Page $\uparrow$
        & \mbox{Exact extr. $\uparrow$} & \mbox{Descriptive ans. $\uparrow$} \\
        \midrule
        \multirow{3}{*}{\shortstack[l]{w/o original image\\at QA time}}
        & MAGI v3$^{\dagger}$~\cite{sachdeva2025panels}
        & 88.89 & 95.18 & 85.21 & 84.51 & 90.91 \\
        & Joint one-step LMM$^{\ddagger}$
        & 88.44 & 91.57 & 86.62 & 84.51 & 90.26 \\
        & MangaParser (Ours)
        & \textbf{92.44} & \textbf{97.59} & \textbf{89.44} & \textbf{90.14} & \textbf{93.51} \\
        \midrule
        \multirow{4}{*}{\shortstack[l]{w/ original image\\at QA time}}
        & MangaLMM~\cite{baek2026mangavqa}
        & 49.33 & 60.24 & 42.96 & 54.93 & 46.75 \\
        & \quad + MangaParser memory
        & 59.11 & 67.47 & 54.23 & 57.75 & 59.74 \\
        & LMM-based VQA
        & 95.56 & 96.39 & 95.07 & 92.96 & 96.75 \\
        & \quad + MangaParser memory
        & \textbf{97.33} & \textbf{98.80} & \textbf{96.48} & \textbf{95.77} & \textbf{98.05} \\
        \bottomrule
    \end{tabular}
    \endgroup
\end{table*}

MangaParser achieves 92.44\% accuracy from its parsing outputs alone, outperforming the joint one-step LMM and MAGI v3 overall and in all four question categories.
MangaParser memory also improves image-based QA for both MangaLMM and LMM-based VQA in every category.

Across all methods, Page questions requiring information from multiple panels have lower accuracy than Panel questions, which can be answered within a single panel.
For Page questions, LMM-based VQA with MangaParser memory achieves 96.48\% accuracy with the original images, compared with 89.44\% from parsing outputs alone.
This gap highlights the value of original visual evidence for reasoning across panels.

By answer type, all three parsers support higher QA accuracy on Descriptive answering than on Exact extraction.
MangaParser improves both answer types, with a larger gain over the joint one-step baseline on Exact extraction (5.63 versus 3.25 percentage points).
These results highlight the value of MangaParser's outputs for retrieving specific information in downstream QA.

\section{Story QA Examples}
\label{sec:supp_qa_examples}

Figure~\ref{fig:supp_story_qa_examples} presents additional Story QA examples using GPT-series backbones.
We show results with and without access to the original images at QA time.
Red and green backgrounds indicate incorrect and correct answers, respectively.

\newlength{\suppqaimageheight}
\newsavebox{\suppqarightpanel}
\begin{figure*}[!t]
    \centering
    \sbox{\suppqarightpanel}{\includegraphics[width=0.49\textwidth]{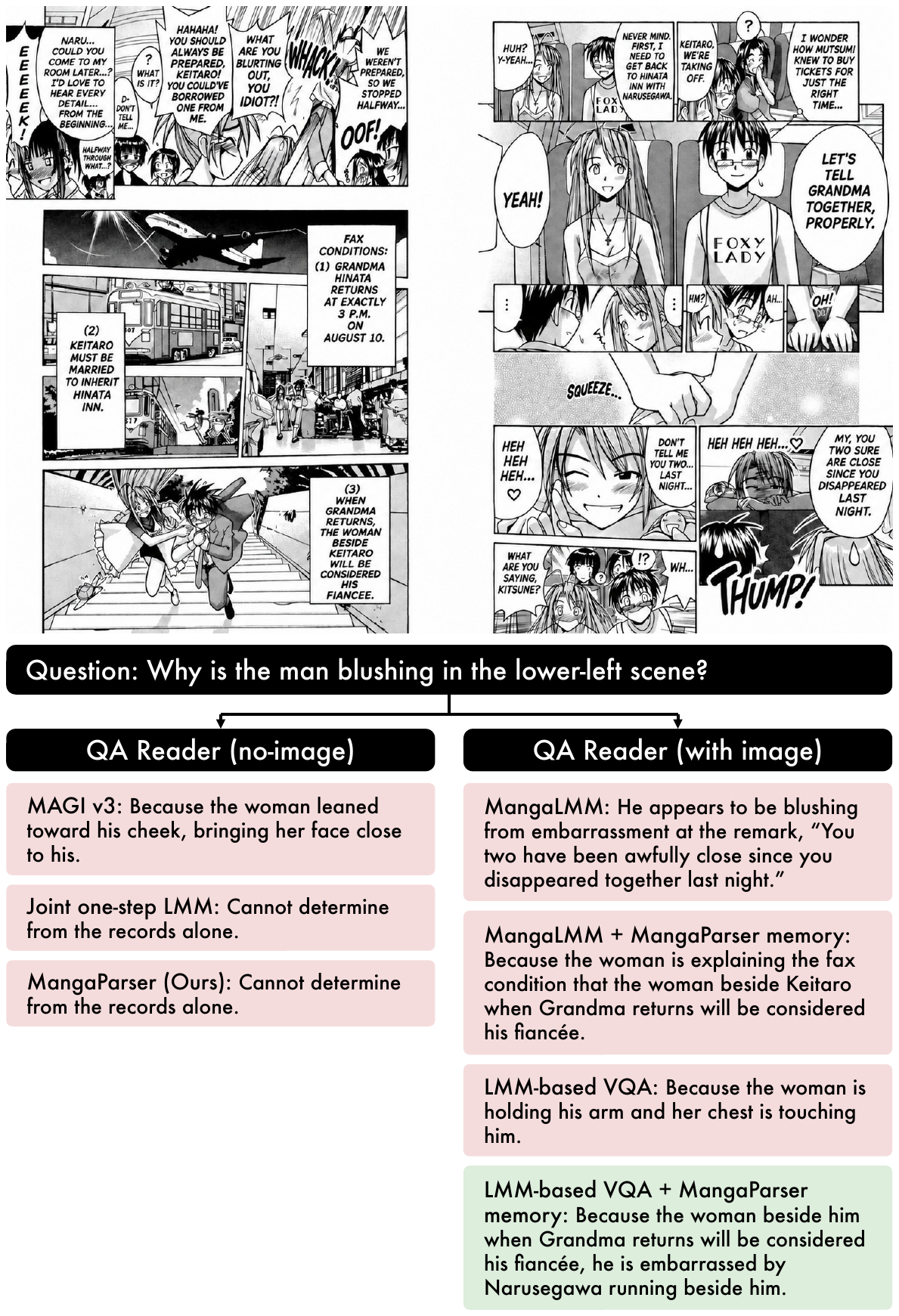}}
    \setlength{\suppqaimageheight}{\ht\suppqarightpanel}
    \begin{subfigure}[t]{0.49\textwidth}
        \vspace{0pt}
        \centering
        \begin{minipage}[t][\suppqaimageheight][t]{\linewidth}
            \vspace{0pt}
            \includegraphics[width=\linewidth]{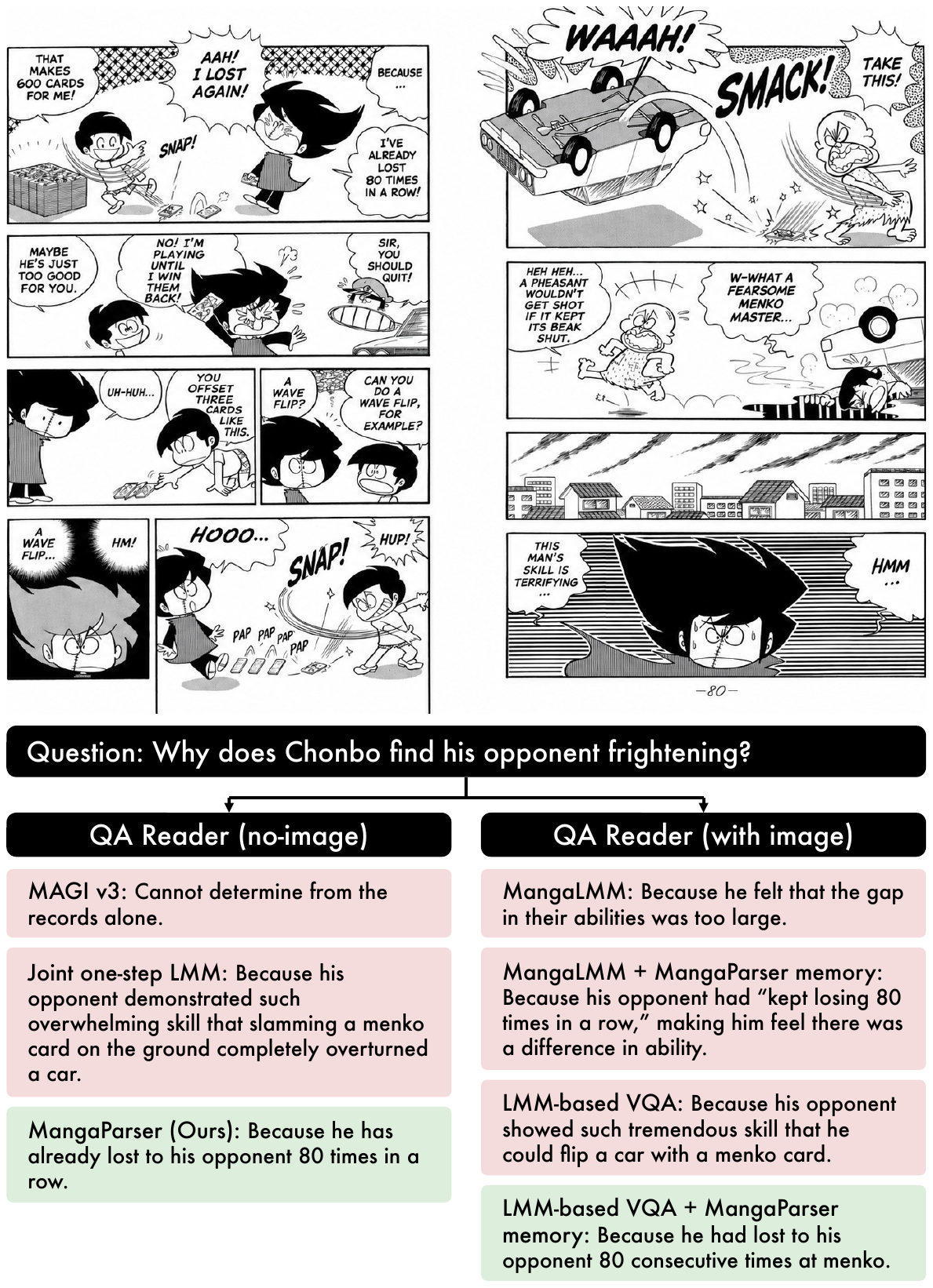}\par
            \vfill
            \nointerlineskip
            \includegraphics[width=\linewidth]{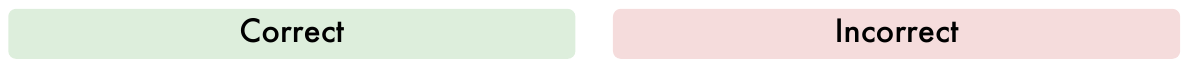}\par
        \end{minipage}
        \caption{Courtesy of Minamoto Taro.}
        \label{fig:supp_qa_chonbo}
    \end{subfigure}\hfill
    \begin{subfigure}[t]{0.49\textwidth}
        \vspace{0pt}
        \centering
        \begin{minipage}[t][\suppqaimageheight][t]{\linewidth}
            \vspace{0pt}
            \usebox{\suppqarightpanel}
        \end{minipage}
        \caption{Courtesy of Akamatsu Ken.}
        \label{fig:supp_qa_lovehina}
    \end{subfigure}
    \caption{Story QA examples using GPT-series backbones.
    Red and green backgrounds indicate incorrect and correct answers, respectively.
    The manga images and QA text are translated from Japanese into English for presentation.}
    \label{fig:supp_story_qa_examples}
\end{figure*}

Figure~\ref{fig:supp_story_qa_examples}~(a) requires extracting the answer from dialogue across panels.
Both the shared text-only QA reader using MangaParser's outputs and LMM-based VQA augmented with its memory answer correctly.
These results demonstrate the effectiveness of our method for Story QA.
With MangaParser memory, MangaLMM also identifies the relevant dialogue, but incorrectly attributes the 80 consecutive losses to the opponent.

Figure~\ref{fig:supp_story_qa_examples}~(b) is more challenging because answering requires locating the relevant panel, recognizing the young man's blush, and connecting his reaction to the surrounding dialogue.
Only LMM-based VQA augmented with MangaParser memory answers correctly.
With MangaParser memory, MangaLMM also identifies the relevant marriage condition, but incorrectly states that the woman explains it.
In future work, we will enrich structured story memory with finer visual details and clearer records of who performs each action.

\end{document}